\documentclass[conference,a4paper]{IEEEtran} \IEEEoverridecommandlockouts
\usepackage{cite}
\usepackage{amsmath,amssymb,amsfonts}
\usepackage{algorithmic}
\usepackage{graphicx}
\usepackage{textcomp}
\usepackage{xcolor}
\def\BibTeX{{\rm B\kern-.05em{\sc i\kern-.025em b}\kern-.08em
    T\kern-.1667em\lower.7ex\hbox{E}\kern-.125emX}}
\begin{document}

\title{Emphasis on the Minimization of False Negatives or False Positives in Binary Classification\\}

\author{Sanskriti Singh$^{1}$}

\maketitle

\begin{abstract}
The minimization of specific cases in binary classification, such as false negatives or false positives, grows increasingly important as humans begin to implement more machine learning into current products. While there are a few methods to put a bias towards the reduction of specific cases, these methods aren't very effective, hence their minimal use in models. To this end, a new method is introduced to reduce the False Negatives or False positives without drastically changing the overall performance or F1 score of the model. This method involving the careful change to the real value of the input after pre-training the model. Presenting the results of this method being applied on various datasets, some being more complex than others. Through experimentation on multiple model architectures on these datasets, the best model was found. In all the models, an increase in the recall or precision, minimization of False Negatives or False Positives respectively, was shown without a large drop in F1 score.
\end{abstract}

\section{Introduction}
As machine learning is incorporated in more fields, the more specific models have to become towards their certain task. One such field is healthcare, where the minimization of False Negatives is far greater than the minimization of False Positives when machine learning is used to diagnose patients [4].  Similarly in the crime department false positives are more dangerous than false negatives since an innocent man could be charged of a crime they didn't do [1]. To create an optimal model that will be of most use to the corresponding field, it should have biases towards cases that are more dangerous than not. For this reason, the creation of a method that reduced specific cases without ruining the overall performance of the model emerged.

To see if the method worked, it was tested on multiple model architectures. The model architectures ranged from linear regression to deep convolutional neural networks. 

Current methods that are available to minimize cases like false negatives include weight change, performing data augmentation to create a biased dataset, and changing the decision boundary line [2]. Many of these methods account for the minimization of a certain case, but fail when it comes to not changing the overall performance. Performing weight change is a common technique used in many models to get a bias towards a certain class, but it kept small in order to not overflow the model to one side. This is performed by making the loss drop lower when that case (ex. FN) is incorrectly detected compared to that of the other case (ex. FP), so the model pays more attention to reducing those cases (ex. FN). Other methods like data augmentation simply involve getting or creating more data of the case you want the model to predict more accurately on. The decision boundary line change is another common method which simply involves changing the y\_hat prediction line from 0.5 to above (reduction of FP) or below (reduction of FN). 

However, examining these methods and learn how effective they are, it can be noticed that they do not prove to be much helpful at extreme levels. They are able to change the recall or precision from 1-2\% at most. A new method was devised to increase the recall metric by around 13\% without losing the F1 score (metric used to measure overall performance of model). This method worked across multiple architectures. 

In this paper, a pneumonia data set is used as a comparison for the performance of the model. Pneumonia is one of the deadliest diseases in the world and is a challenge for both models and humans to detect from chest x-rays. The reasons for the difficulty of detection of pneumonia include its increasing similarity to other lung diseases such as bronchitis, its faint opacity in a completely black and white image, and more. 

\section{Data}

\begin{figure}[htp]
\centerline{\includegraphics[width=9cm]{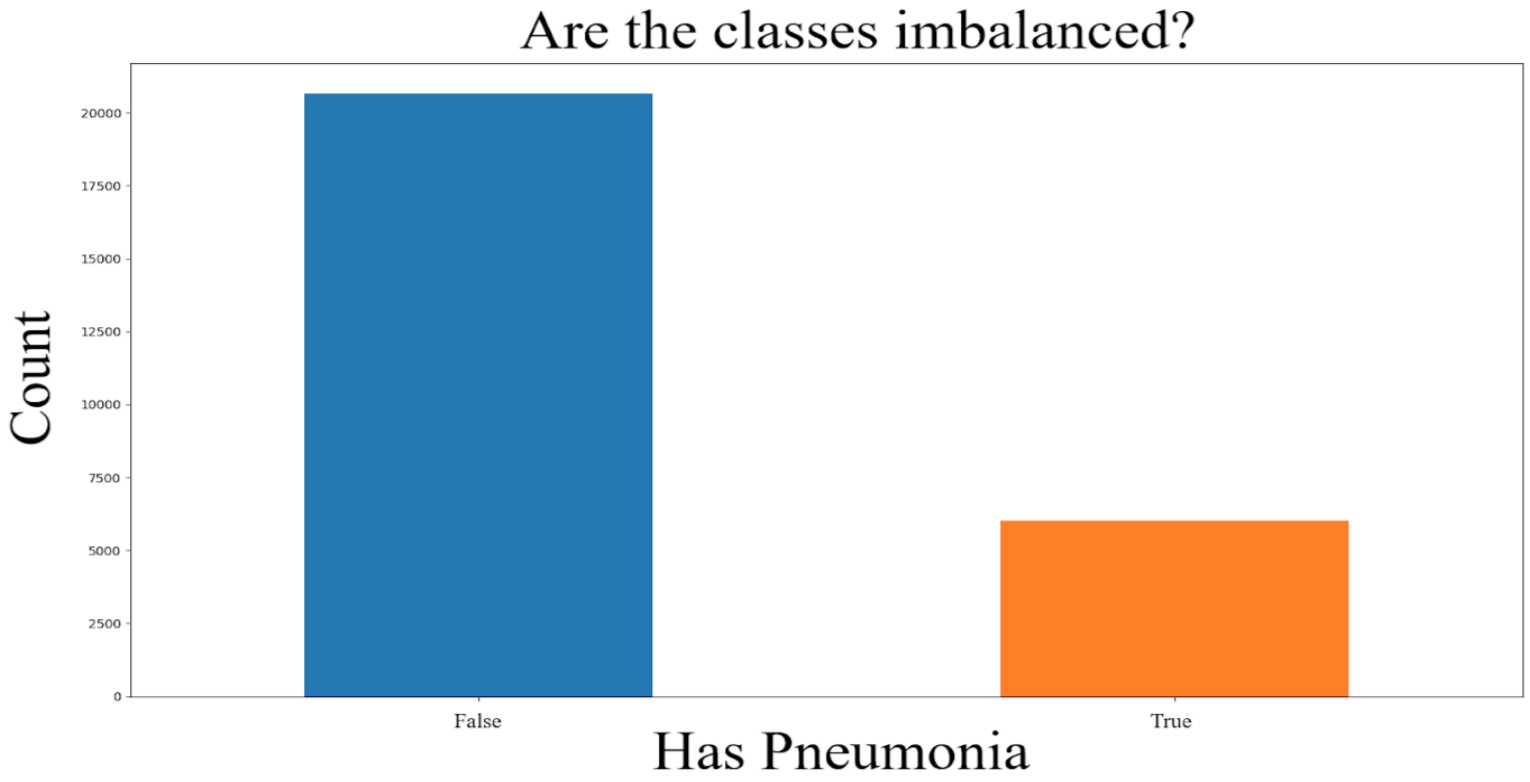}}
\caption{An imbalanced dataset with a large ratio of no pneumonia to pneumonia images; fixed through augumentation.}
\label{fig:p}
\end{figure}

The main dataset used was the RSNA dataset provided by the Radiological Society of North America, containing 26,684 frontal X-ray images split into the categories of pneumonia and no pneumonia [5]. Augmentation was performed to balance the images with no pneumonia versus ones with pneumonia. Horizontal and/or Vertical augmentation was performed on the true dataset to create a total of 33,463 images. (Figure \ref{fig:p}) This dataset was split into 80\% train and 20\% test. After data augmentation on the training dataset, it contained 33463 images. This augmented dataset was further split into 80\% train and 20\% validation.

\begin{figure}[htp]
\centerline{\includegraphics[width=9cm]{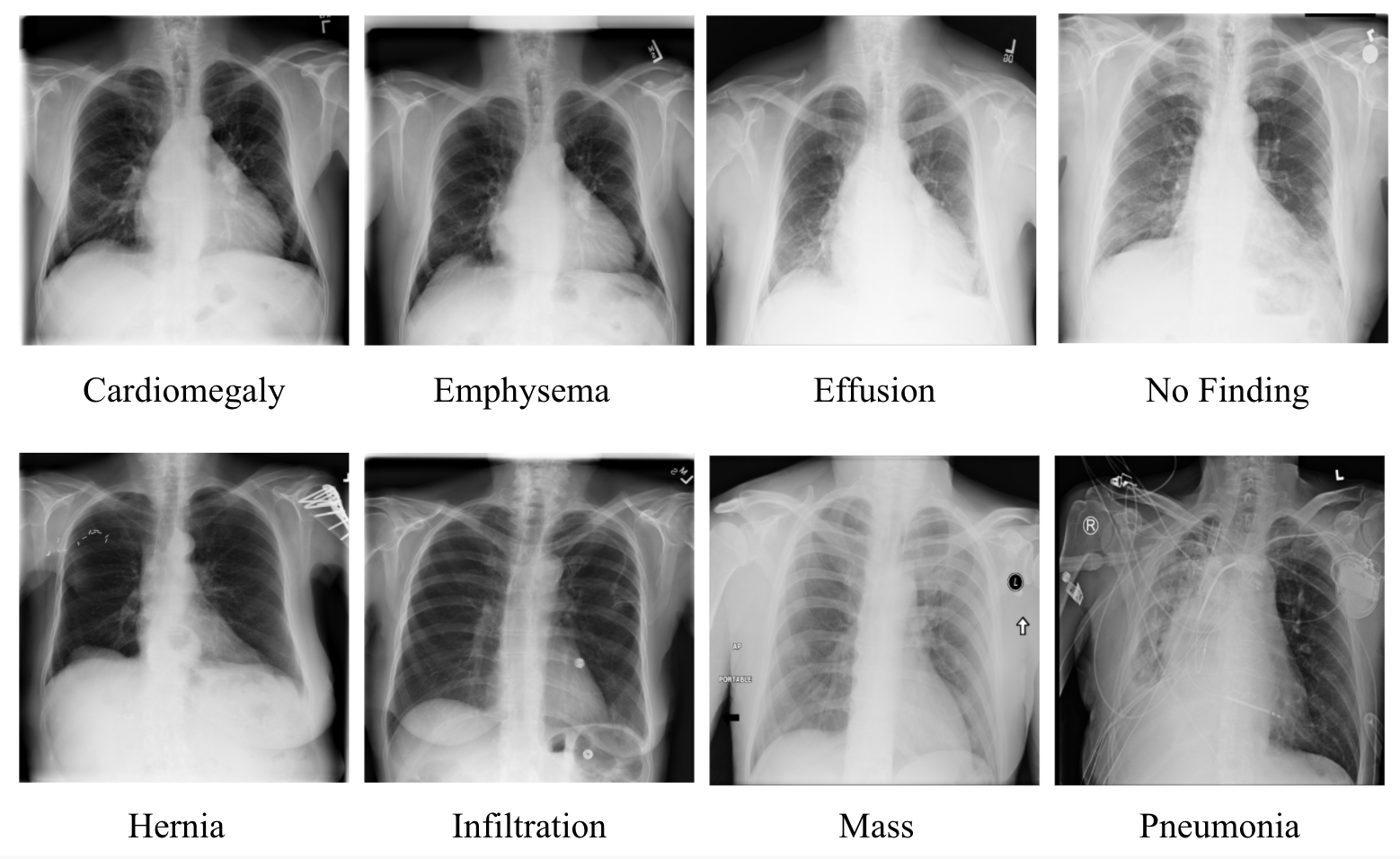}}
\caption{An overview as to what changes are done to the data before re-trained. Some percent of the False Positives are taken to change their real value to true therefore becoming True Positives.}
\label{fig:p1}
\end{figure}

Along with this pneumonia dataset, many models were trained through transfer learning on the infamous NIH chest, ChestX-ray8, containing 141665 frontal x-ray images classified into 8 different dieases, including pneumonia [7]. This dataset was not used on its own, but used to perform a transfer-learning model which also was tested on the false negative method (Figure \ref{fig:p1}).

\section{Function}

The methodology behind the proposed method of minimizing the number of False Negatives without ruining the entire performance of the model involves data alteration. Rooted from the combination of data augmentation and threshold change, a pre-trained model was taken and re-trained on the same set of data with slightly different real values. Data augmentation is where you try and create more data for the model to train on so it performs better since it has more resources to gather info from. Threshold change is where you change the y\_hat decision boundary line from 0.5 to something above or below. In the case of minimization of False Negatives, optimally the threshold would want to go below 0.5. This method works but creates problems if implemented with testing as it forms a unwanted bias. This method alters a portion of the data in order to learn to change the decision boundary line during training without forcing a model to think a certain way. 

\begin{figure}[htp]
\centerline{\includegraphics[width=9cm]{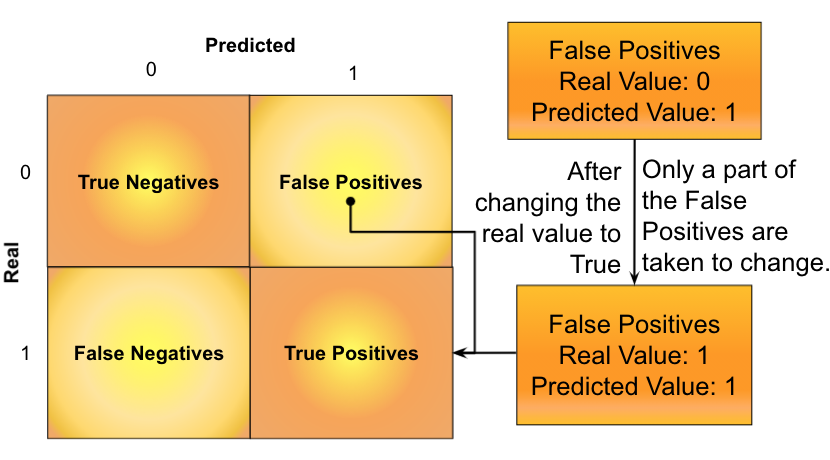}}
\caption{An overview as to what changes are done to the data before re-trained. Some percent of the False Positives are taken to change their real value to true therefore becoming True Positives.}
\label{fig:pneumoxttention}
\end{figure}

Taking the training set and predicting on it with the pre-trained model, produced a 2x2 confusion matrix of true negatives, false negatives, true positives, and false positives [3]. The images that fell under the false positive category were changed so that their real value was 1. To see how effective this method truly was experimentation was done on different portions of the false positive images (Figure \ref{fig:pneumoxttention}). 

\begin{figure}[htp]
\centerline{\includegraphics[width=9cm]{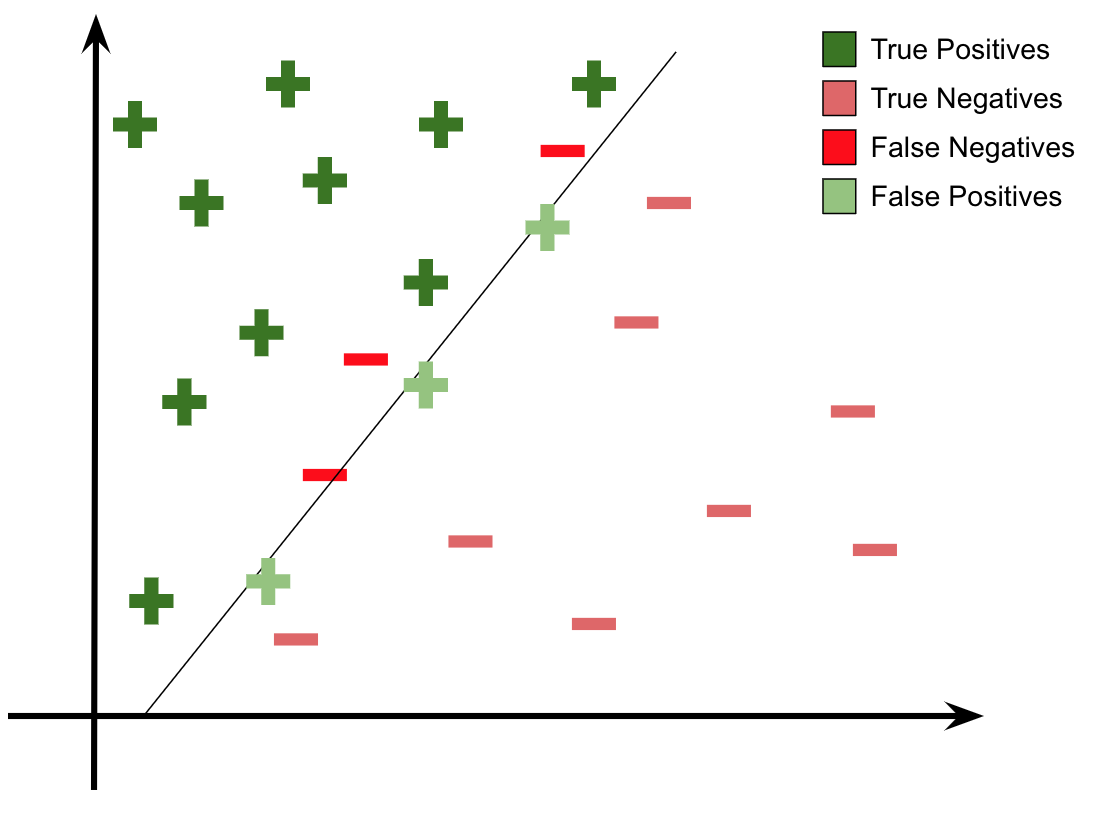}}
\caption{A 2d representation of the predicted data values.}
\label{fig:p2}
\end{figure}

This works because some of the False Positives predicted have similar features to some of the False Negatives, hence when changing the model to train on the False Positives with the notion that these are the "true" features, it learns to identify with a bias towards images with pneumonia. In (Figure \ref{fig:p2}) it can be visualized that many of the incorrect cases lie close to each other, meaning the model sees similar features between the two. Training the model to see these features as true, or having pneumonia it will put an emphasis on the minimization of False Negatives. The decision boundary line will theoretically move to the left. Vice Versa would be performed if it wanted to reduce False Positives. 

This method can be compared to the threshold line change which performs similarly. The difference between the two methods is that the model continues to learn to perfect its decision boundary line to fit the new data points, whereas the threshold line change hard codes the change so its predictions are less learned and more forced.

\section{Architecture}

\begin{figure}[htp]
\centerline{\includegraphics[width=9cm]{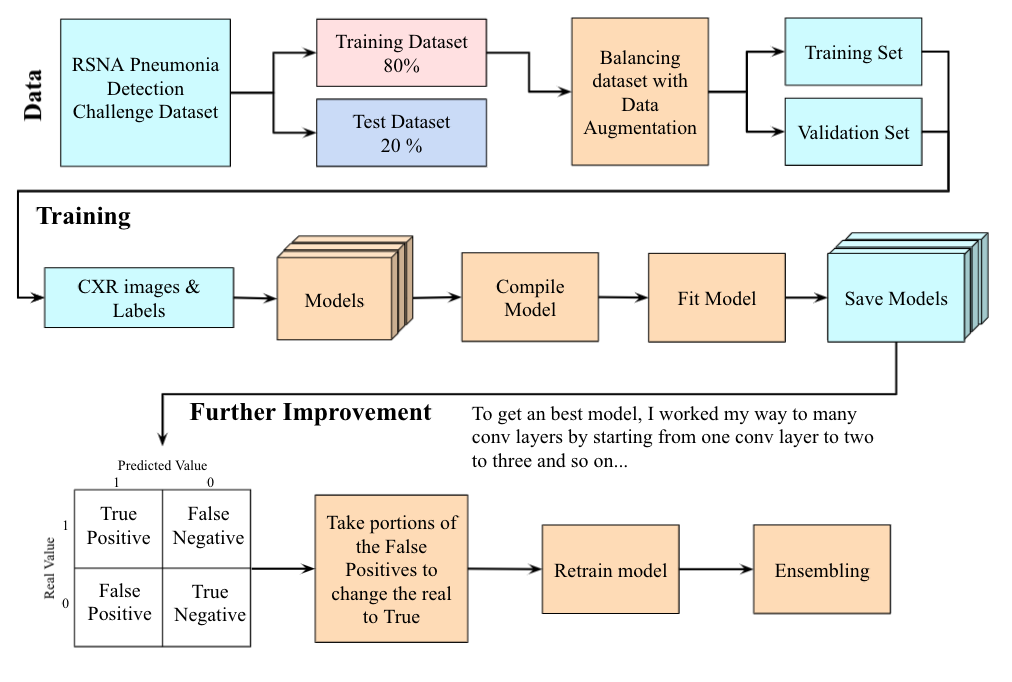}}
\caption{Model Flowchart}
\label{fig:qwe}
\end{figure}

Since the goal was to test the efficiency of the method involved to put an emphasis on the minimization of False Negatives, the architecture of the entire process was kept comparatively simple (Figure \ref{fig:qwe}). Instead, focusing more on making complex models to test the method on which deep architectures as well as famous architectures.

\begin{figure}[htp]
\centerline{\includegraphics[width=9cm]{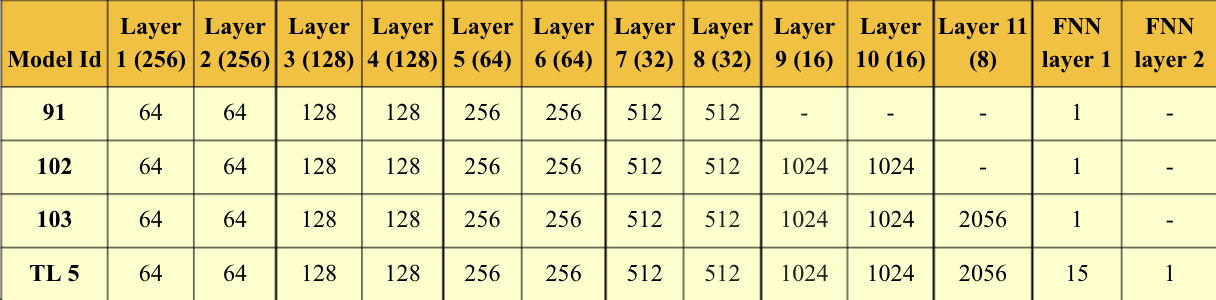}}
\caption{Final Base Models' Architectures}
\label{fig:p3}
\end{figure}

To test the method's affect on the performance of the model it was optimal to reach high performing models before applying the method. The search began by taking the best model architecture with one layer, then two, then three, and on and on. the best model was then chosen through its results on the validation set. The test set was completely untouched throughout the entire process. After this process, 5 best models were chosen in which the method was soon applied. 3 of these models came from this experimentation process, and the other two included the infamous vgg16 model architecture and a transfer learning model [6]. The transfer learning model was trained to create a deeper, more complex model. It involved training the NIH dataset on a 11 layer CNN model and retraining the model onto the binary RSNA dataset through one hidden layer of 15 neurons. These models were chosen based on their performance and complexity to test the range and value of the minimization of False Negatives method devised. The first 3 models were deep neural network architectures that performed phenomenally well on the validation set. VGG16 also performed extremely well and is also a very famous/complex model. The transfer learning model was chosen for its excellent results on the validation data as well as its complexity. (Figure \ref{fig:p3})

\section{Comparison}

\subsection{Class Weights}

A famous method often used to create a model with an emphasis on the minimization of false negatives is changing the class weights. This was originally introduced as a solution for imbalanced data sets, to balance the under representation of a specific class. This method was soon recognized for also being able to create a bias towards a certain class.

While the method of class weights decreased the number of false negatives, hence the increase in recall, the F1 score was inconsistent and often decreased (table I). Comparing this to the proposed method, it was comparatively worse since this method increased recall and mostly kept the F1 score.

\begin{table}[h!]
\centering
\caption{\label{table:1}Minimization of FN using Class Weights on Model 91}
\begin{tabular}{ |c|c|c|c|c| } 
 \hline
 \textbf{Class Weight} & \textbf{Recall} & \textbf{Precision} & \textbf{F1 score} & \textbf{AUROC} \\ 
 \hline
 0:1, 1:50 & 0.98 & 0.26 & 0.41 & 0.61\\
 \hline
 0:1, 1:25 & 0.83 & 0.4 & 0.54 & 0.75\\
 \hline
 0:1, 1:10 & 0.64 & 0.41 & 0.52 & 0.7\\
 \hline
 0:1, 1:2 & 0.57 & 0.53 & 0.55 & 0.72\\
 \hline
 0:1, 1:1 & 0.53 & 0.59 & 0.56 & 0.71\\
 \hline
\end{tabular} 
\end{table}

Comparing this method of class weight change to the base model with a recall of 0.53 and F1 score of 0.56, it is seen that recall has a consistency of increasing at the large cost of the F1 score or overall performance of the model. The reasoning behind these results is that the loss function favors the positive cases over the negative cases meaning that when a negative image (without pneumonia) is misdiagnosed it has a small effect on the loss so the model continues to learn without paying too much attention towards the decreasing precision. With a larger class weight the recall seems to increase in proportion to the decrease of precision. An extreme example of this is that one error in a positive case (with pneumonia) is the same as 50 error in a negative case (without pneumonia) so the model inherently favors the positive cases. The problem with this method is that there is no way to increase the recall without dropping the precision so much as to ruining the overall performance of the model. 

\subsection{Threshold Line}

Another common method used to decrease cases like false negatives or false positives is changing the decision boundary line. The basic decision boundary line in binary classification models is 0.5. When the \^{y} value is greater than 0.5, the prediction is considered True. When the \^{y} value is lower than 0.5, the prediction is False.

To reduce False Negatives the threshold line was lowered since that would force the model to predict less inputs as False, therefore reducing the number of False Negative cases. Similarly increasing the threshold line decreased the number of False Positives.

Like the class weights method, while this lowers False Negatives or False Positives, it comes at the cost of increasing the opposite false case (table II). While the reduction of the number of False Negatives was possible, the number of False Positives increased dramatically. 

\begin{table}[h!]
\centering
\caption{\label{table:1}Minimization of FN through Threshold Change on Model 91}
\begin{tabular}{ |c|c|c|c|c| } 
 \hline
 \textbf{Threshold Line} & \textbf{Recall} & \textbf{Precision} & \textbf{F1 score} \\ 
 \hline
 0.0 & 1.00 & 0.00 & 0.56\\
 \hline
 0.1 & 0.98 & 0.26 & 0.41\\
 \hline
 0.2 & 0.54 & 0.4 & 0.54\\
 \hline
 0.3 & 0.54 & 0.4 & 0.52\\
 \hline
 0.4 & 0.53 & 0.59 & 0.56\\
 \hline
 0.5 & 0.53 & 0.59 & 0.56\\ 
 \hline
\end{tabular} 
\end{table}

Comparing this method of threshold change to the base model with a recall of 0.53 and F1 score of 0.56, it is seen that unlike the method of class weight change and the method proposed, this isn't a learned method.  

\section{Results}

When applying the proposed method to the 5 different base models experimentation was done to try and change the real value of different amounts of the False Positives (table III). It can be noticed that at 20\% and 40\% it was more unlikely to see a large change in recall or minimization of False Negatives. At around 60\% and 80\% change of the real value of false positives there was a visible chnage and the F1 score usually stayed consistent. At 100\% change it was a bit more unpredictable given that all of the False Positives real value is changed into 1. Overall the models showed an emphasis on the minimization of False Negatives with the same overall performance or F1 score.

The affect of this method is very dependent on the base model. It is noticed that when the base model has a higher precision (91, 102, 103), the affect of this method is more visible as the recall increases by 1-5\%. Sometimes the precision also increases, since the model is still learning and isn't' forced to predict a certain way. In base models where the precision starts out lower such as transfer learning and VGG16 the method's effectiveness is less visible. A possible reason behind this is that with a higher precision there are less false positives for the method to be applied to. This method of minimizing false negatives relies entirely on how the false positives are chosen for changing the real value of. If the number of false positives starts out as small, the false positives that are available for change are more concentrated, less likely to be outliers. With a low precision, there is a lot of false positives. This makes matters of choosing False Positives more difficult since the 10\% you choose could be very different from each other, making no difference on the learning of the model. With large percentages like 80\% and 100\% you are taking a large amount of the data and changing its real value which could potentially mean a different dataset. With this method the amount of data you choose to change is vital to its proper effectiveness in the real world.

\begin{table}[h!]
\centering
\caption{\label{table:1}Minimization of False Negatives}
\begin{tabular}{ |c|c|c|c|c| } 
 \hline
 \textbf{Model} & \textbf{Percentage of Change} & \textbf{Recall} & \textbf{Precision} & \textbf{F1 score} \\ 
 \hline
 91 & 0\% & 0.53 & 0.59 & 0.56 \\
 \hline
 91 & 20\% & 0.58 & 0.54 & 0.56\\
 \hline
 91 & 40\% & 0.64 & 0.5 & 0.56\\
 \hline
 91 & 60\% & 0.53 & 0.56 & 0.55\\
 \hline
 91 & 80\% & 0.58 & 0.54 & 0.56\\
 \hline
 91 & 100\% & 0.58 & 0.54 & 0.56\\
 \hline
 102 & 0\% & 0.5 & 0.55 & 0.53 \\
 \hline
 102 & 20\% & 0.49 & 0.54 & 0.51\\
 \hline
 102 & 40\% & 0.49 & 0.54 & 0.51\\
 \hline
 102 & 60\% & 0.54 & 0.54 & 0.54\\
 \hline
 102 & 80\% & 0.51 & 0.55 & 0.53\\
 \hline
 102 & 100\% & 0.52 & 0.56 & 0.54\\
 \hline
 103 & 0\% & 0.55 & 0.51 & 0.53\\
 \hline
 103 & 20\% & 0.55 & 0.51 & 0.53\\
 \hline
 103 & 40\% & 0.56 & 0.51 & 0.54\\
 \hline
 103 & 60\% & 0.56 & 0.51 & 0.55\\
 \hline
 103 & 80\% & 0.56 & 0.55 & 0.56\\
 \hline
 103 & 100\% & 0.56 & 0.54 & 0.56\\
 \hline
 transfer & 0\% & 0.62 & 0.49 & 0.54 \\
 \hline
 transfer & 20\% & 0.63 & 0.5 & 0.56\\
 \hline
 transfer & 40\% & 0.55 & 0.55 & 0.55\\
 \hline
 transfer & 60\% & 0.6 & 0.53 & 0.56\\
 \hline
 transfer & 80\% & 0.66 & 0.48 & 0.57\\
 \hline
 transfer & 100\% & 0.7 & 0.48 & 0.57\\
 \hline
 VGG16 & 0\% & 0.74 & 0.33 & 0.45 \\
 \hline
 VGG16 & 20\% & 0.73 & 0.33 & 0.45\\
 \hline
 VGG16 & 40\% & 0.74 & 0.32 & 0.45\\
 \hline
 VGG16 & 60\% & 0.78 & 0.32 & 0.48\\
 \hline
 VGG16 & 80\% & 0.7 & 0.35 & 0.47\\
 \hline
 VGG16 & 100\% & 0.75 & 0.33 & 0.46\\
 \hline
\end{tabular} 
\end{table}

To see if these models really improved the recall while staying near the same F1 score, ensembling of the best models was performed without the emphasis on the minimization of False Negatives and compared it to another ensembled model with the false negatives minimized (table IV).

\begin{table}[h!]
\centering
\caption{\label{table:1}Normal Vs. False Negatives Minimization}
\begin{tabular}{ |c|c|c|c|c| } 
 \hline
 \textbf{Model} & \textbf{Recall} & \textbf{Precision} & \textbf{F1 score} & \textbf{AUROC}\\ 
 \hline
 Before: & 0.56 & 0.6 & 0.58 & 0.73 \\
 \hline
 After: & 0.68 & 0.51 & 0.58 & 0.76\\
 \hline
\end{tabular} 
\end{table}

\section{Discussion}

In the data science community there have been many implementations relating to the medical machine learning field. These include improving the model's overall performance or creating methods to create a beneficial bias for real time usage. Diagnosis through chest radiograph x rays is quite popular in this field. Many improvements to this has been done by creating deeper neural networks, devising new ways to minimize certain cases over others, and adding new methods to benefit overall performance of the model.

The idea of minimizing a certain case over another is popular since the growing inclusion of AI in the real world needs biases to survive and creater a safer community. 

Some of these methods include class weight change from keras documentation, data augumentation, and threshold change.

Some further research in this topic involve testing this method on False Positive minimization and using deeper neural networks. While the method was tested on famous networks like VGG16, it can be applied to ever deeper neural networks on larger datasets. By trying this method on different datasets, the effectiveness can be analyzed and shown on different types of data. 

\section{Conclusion}

Machine Learning technologies are being developed at a fast rate. As they are applied to different fields its important for them to be molded to the situation. In most medically related cases, the false negative case is highly dangerous in comparison to the false positive. By the creation of this method, the model diagnosing pneumonia or another possible disease would keep this danger risk in mind when making its final prediction, possibly saving more lives. This model would be kept as a second hand tool in hard decision making tasks such as this one. By improving the recall by around 13\% and keeping the F1 score in a complex detection task, it proves its effectiveness. 



\end{document}